\pdfoutput=1

\documentclass[11pt]{article}
\usepackage{EMNLP2023}
\usepackage{times}
\usepackage{latexsym}

\usepackage[T1]{fontenc}
\usepackage[utf8]{inputenc}
\usepackage{microtype}
\usepackage{inconsolata}

\usepackage{amsthm}

\usepackage{amsmath}
\usepackage{graphicx}
\usepackage{algorithm}
\usepackage[noend]{algpseudocode}

\usepackage{enumitem}

\usepackage{multirow}
\usepackage{adjustbox}
\usepackage{bbm}
\usepackage{wrapfig,lipsum}
\usepackage{xspace}
\usepackage{booktabs,cellspace}  
\usepackage[normalem]{ulem}
\usepackage{makecell}
\usepackage{amssymb}
\usepackage{pifont}
\newcommand{\cmark}{\ding{51}}%
\newcommand{\xmark}{\ding{55}}%

\usepackage{microtype}

\usepackage{cleveref}

\newcommand{\modelname}{\textsc{ENTRE}\xspace}
\newcommand{\dataname}{\textsc{ENTRED}\xspace}

\usepackage{etoolbox}
\usepackage[tikz]{bclogo}

\usepackage{subcaption}
\usepackage{adjustbox}

\definecolor{msftBlue}{RGB}{0,164,239}
\definecolor{msftGreen}{RGB}{127,186,0}
\definecolor{msftYello}{RGB}{255,185,0}
\definecolor{msftBlack}{RGB}{0,0,0}
\newcommand{\finding}[1]{
	\begin{bclogo}[couleur= msftBlue!15,  arrondi=0, logo=\bcquestion, marge=2,   couleurBord=msftBlue!10,  sousTitre ={\em \textit{\textbf{#1}}}]{} 
	\end{bclogo}
	\vspace{-1.4em}
}

\title{How Fragile is Relation Extraction under Entity Replacements?}

\author{
Yiwei Wang$^\dagger$ \ \ \ \ Bryan Hooi$^\ddagger$ \ \ \ \ Fei Wang$^\mathsection$ \ \ \ \ Yujun Cai$^\mathparagraph$\thanks{\;Meta and its employees did not perform the experiments or use or process any of the data described or referenced in the paper.} \ \ \ \  Yuxuan Liang$^\|$ \ \ \ \ \\ \textbf{Wenxuan Zhou$^\mathsection$ \ \ \ \ Jing Tang$^\|$ \ \ \ \ Manjuan Duan$^{\dagger\dagger}$ \ \ \ \ Muhao Chen$^{\ddagger\ddagger}$} \\ 
$^\dagger$ University of California, Los Angeles \quad $^\ddagger$ National University of Singapore \\ 
$^\mathsection$ University of Southern California  \quad
$^\mathparagraph$ Meta \quad \\
$^\|$ Hong Kong University of Science and Technology (Guangzhou) \\
$^{\dagger\dagger}$ Amazon \quad $^{\ddagger\ddagger}$ University of California, Davis \\
\texttt{wangyw\_seu@foxmail.com}
}

\date{}

\begin{document}
\maketitle
\begin{abstract}
Relation extraction (RE) aims to extract the relations between \textbf{entity names} from the \textbf{textual context}.
In principle, textual context determines the ground-truth relation and the RE models should be able to correctly identify the relations reflected by the textual context.
However, existing work has found that the RE models memorize the entity name patterns to make RE predictions while ignoring the textual context.
This motivates us to raise the question: ``are RE models robust to the entity replacements?''
In this work, we operate the random and type-constrained entity replacements over the RE instances in TACRED and evaluate the state-of-the-art RE models under the entity replacements. 
We observe the 30\% - 50\% F1 score drops on the state-of-the-art RE models under entity replacements.
These results suggest that we need more efforts to develop effective RE models robust to entity replacements. 
We release the source code at \url{https://github.com/wangywUST/RobustRE}.
\end{abstract}

\section{Introduction}\label{sec:int}
Recent literature has shown that the sentence-level relation extraction (RE) models may overly rely on entity names for RE instead of reasoning from the textual context \cite{peng2020learning,wang2022should}.
This problem is also known as \emph{entity bias}: the spurious correlation between entity names and relations \cite{longpre2021entity,qian-etal-2021-annotation,xu-etal-2022-model,wang2022should}.
This motivates us to raise a question: ``how robust are RE models under entity replacements?''

\begin{figure}[!tb]
	\centering
	\includegraphics[width=1\linewidth]{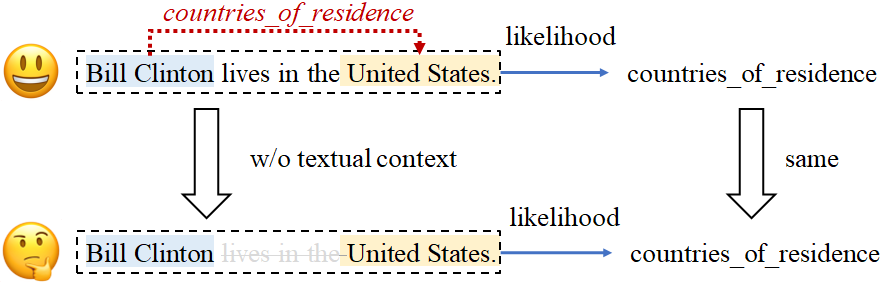}
	\caption{
    TACRED offers many shortcuts from entity names to ground-truth relations in the test set, where the model predicts the correct relation even when only given the entity names, despite all textual context being removed. 
    As a result, TACRED is not challenging enough to measure the generalization under entity bias.
\label{fig:1}}
\end{figure}

Entity bias degrades the RE models' generalization, such that the entity names can mislead the models to make wrong predictions. 
However, a seemingly conflicting phenomenon is the high (in-distribution) accuracy of RE models on the standard benchmarks, such as TACRED.
In our work, we find that these benchmarks are prone to have shortcuts from entity names to ground-truth relations (see Fig.~\ref{fig:1}), low entity diversity, and a large portion of incorrect entity annotations.
These issues suggest that, given the presence of entity bias, the current benchmarks are not challenging enough to evaluate the generalization of RE in practice.

Evaluating RE with valid instances of more comprehensive entities is non-trivial.
It requires us to collect many sentences containing comprehensive entities and carefully label the relations.
Both the text collection and annotations are time-consuming and expensive.
Instead, in our work, we aim to efficiently produce rich valid RE instances with comprehensive entities based on the carefully designed entity replacements.
Most existing methods for evaluating the generalizability of NLP focus on sentence classification \cite{jin2020bert, Li2020BERTATTACKAA, minervini2018adversarially} and question answering \cite{jia2017adversarial, ribeiro-etal-2018-semantically, gan2019improving}, but these methods lack special designs to seize on the entity bias in RE.

\begin{figure}[!tb]
	\centering
	\includegraphics[width=1\linewidth]{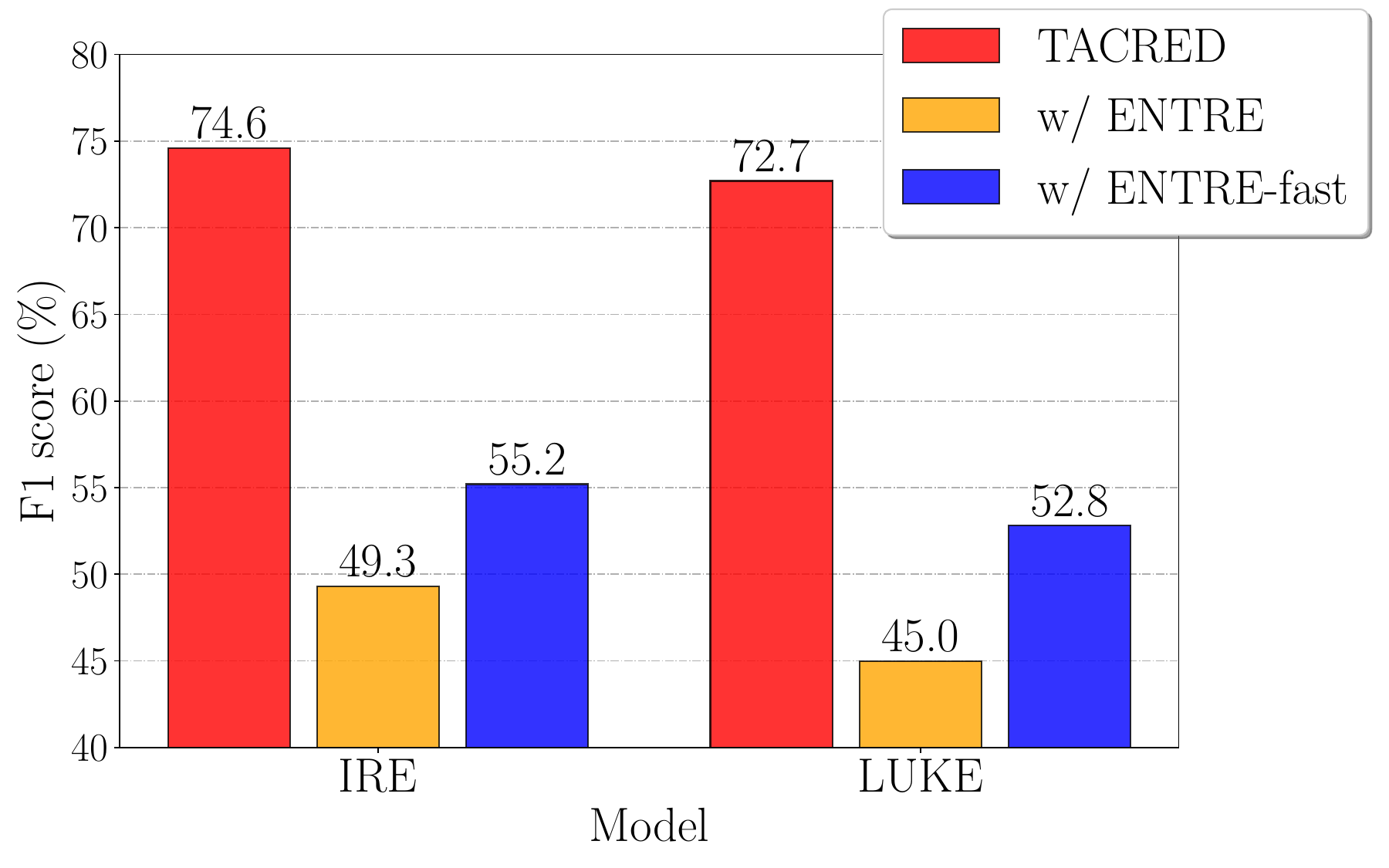}
	\caption{
    The performance of state-of-the-art RE models drop a lot under entity replacements (ENTRE).
\label{fig:replace}}
\end{figure}

In this work, we propose a \textbf{type-constrained} and \textbf{random} entity replacement method: \modelname. 
\textbf{Type-constrained} means we replace the named entity in the type \texttt{[PERSON]} or \texttt{[ORGANIZATION]} with the new entity belonging to the same type as the original entity.
\textbf{Random} means we randomly select the entity names from a Wikipedia entity lexicon that consists of 24,933 organizations and 902,007 person entities for replacements.
These two principles guarantee the effectiveness of entity replacement to produce valid and diverse RE instances.

We apply \modelname to TACRED and evaluate the RE models on the instances with replaced entity names.
We analyze the RE models under entity replacements in order to answer four research questions: (Q1) How do the strong RE models perform under entity replacements?
(Q2) Does \modelname reduce prediction shortcuts from entity names to the ground-truth relations? (Q3) Does \modelname improve the entity diversity? 
(Q4) How to improve the robustness of RE?

We observe several key findings.
First, the strong RE models LUKE \cite{yamada-etal-2020-luke} and IRE \cite{zhou2021improved} tend to memorize entity-relation patterns to infer the relation instead of reasoning based on the textual context that actually describes the relation.
This phenomenon causes the model to be brittle to entity replacements, resulting in a significant performance drop of 30\% - 50\% in terms of the F1 score.
Second, \modelname reduces the shortcuts by more than 50\% on many relations, and improves the subject name diversity by more than 25 times compared to TACRED.
Third, the recent causal inference approach CoRE \cite{wang2022should} improves the robustness at a higher magnitude than other methods.

For the easy use of \modelname, we provide a challenging RE benchmark built by \modelname: \dataname, which consists of the TACRED test set instances with the entity names replaced by \modelname.
We believe the proposed \modelname and benchmark \dataname will benefit future research toward improving the RE robustness.

\section{Analysis of Entity Names in TACRED}

Before introducing \modelname, we first analyze the existing popular RE datasets. 
Our analysis is focused on the following three perspectives: 1) the correctness of entity name annotations; 2) the diversity of entity names; 3) the prediction shortcuts from entity names to the ground-truth relations.

In the popular TACRED \cite{zhang2017position}, TACREV \cite{alt2020tacred}, and Re-TACRED \cite{retacred} datasets, we find that: first, there exist some portion of incorrect entity name annotations; second, many entity names are reused more than one hundred times across instances; third, the entity names in more than 70\% of the instances act as shortcuts to the ground-truth relations.
We introduce the details as follows.

\subsection{Incorrect Entity Annotations} \label{sec:2_1}
\begin{figure}[!tb]
	\centering
 \includegraphics[width=1\linewidth]{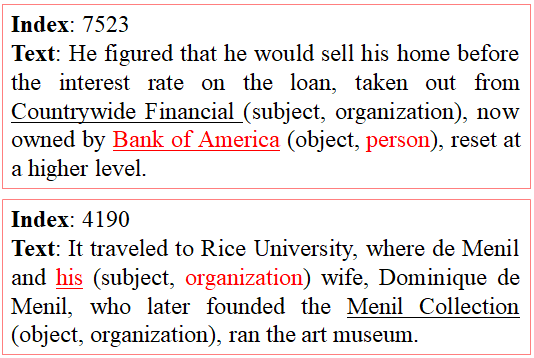}
	\caption{
	Two examples of incorrect entity annotations in TACRED. \label{fig:annot}}
\end{figure}

In the TACRED \cite{zhang2017position}, TACREV \cite{alt2020tacred}, and Re-TACRED \cite{retacred} datasets, there exist quite a few incorrect entity annotations. 
To detect these incorrect entity annotations, we use a BERT based NER model \cite{devlin-etal-2019-bert} to automatically annotate the subject and object entity names in the TACRED dataset.
Then, we conduct 
manual investigation on the entities where the NER annotations are different the original TACRED annotations.
We find that more than 10\% of the test instances contain incorrect entity annotations.\footnote{Including both incorrect span and type annotations.}
We present two examples in Fig.~\ref{fig:annot}.
Using these mistaken entity annotations to evaluate the RE models compromises our goal of correctly measuring RE performance.

\begin{figure}[!tb]
	\centering
	\includegraphics[width=1\linewidth]{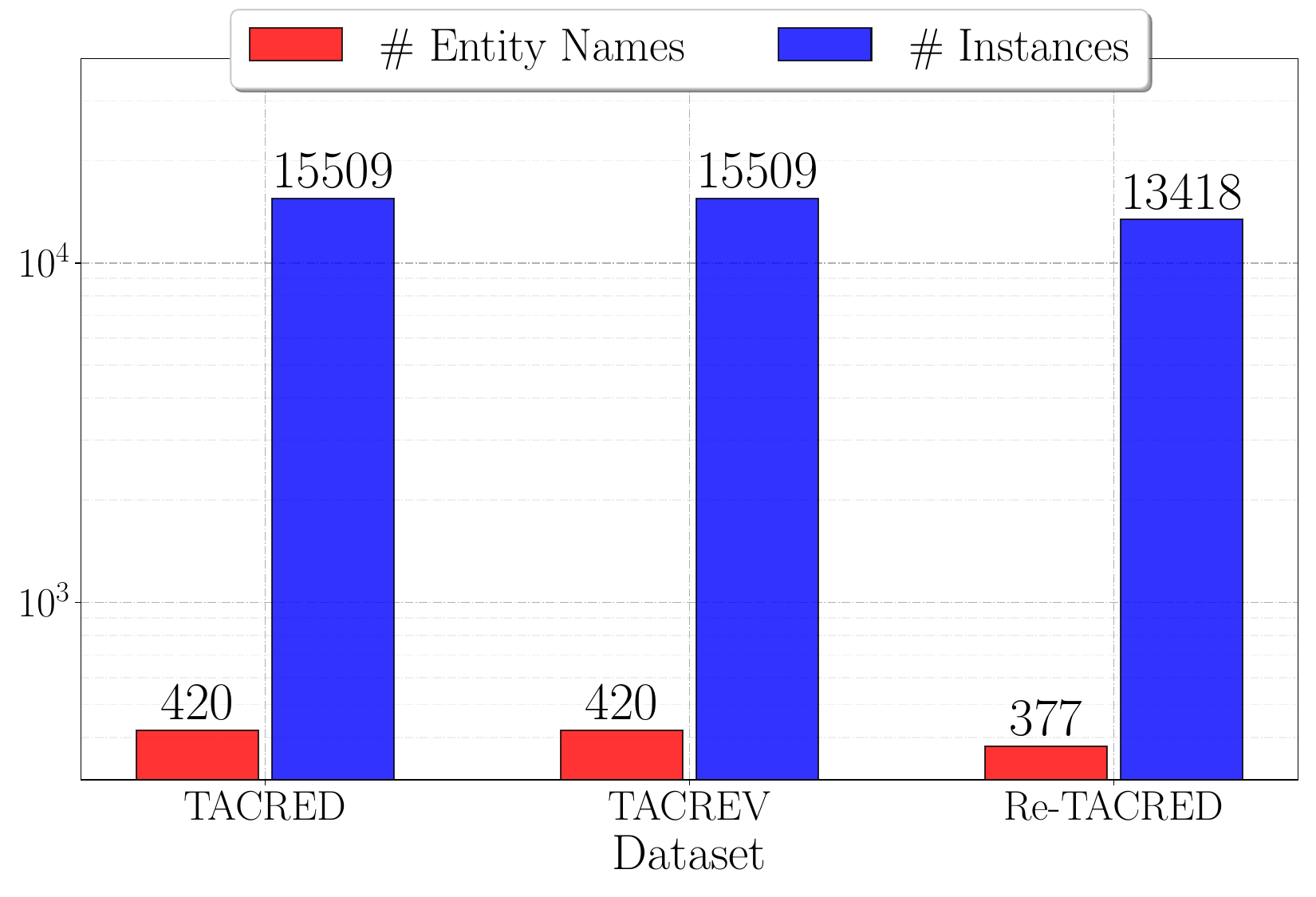}
	\caption{
	The number of different subject entity names (red) is much lower than the number of instances (blue) in the test sets of the TACRED, TACREV, and Re-TACRED datasets.
        In other words, the diversity of entity names in these datasets' test sets is limited.
		\label{fig:number_name}}
\end{figure}

\subsection{Diversity of Entity Names}
The TACRED, TACREV, and Re-TACRED datasets have a low diversity of entity names: most entity names repeatedly appear in 
a large portion of instances  (see Fig.~\ref{fig:number_name}). 
In the TACRED datasets, there are only 420 entity names repeatedly appearing as 15509 instances' subjects.
For example, \textit{``ShopperTrak''}, as the subject, has repeatedly appeared as the subject entity in 270 instances.
This heavily repeated use of entity names increases the risk that RE relies on entity bias to make RE predictions.
Also, with these benchmarks, it is impossible to comprehensively evaluate the generalization of RE models on a diverse set of entity names to imitate real-world scenarios.

\subsection{Causal Inference for Entity Bias}\label{sec:3_1}
\begin{figure}[!tb]
	\centering
	\includegraphics[width=1\linewidth]{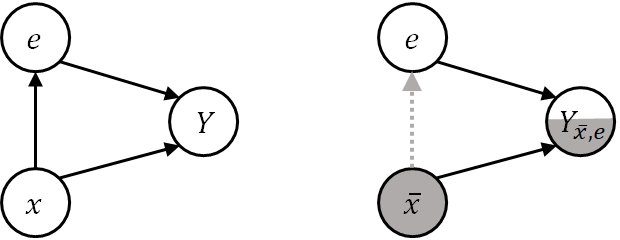}
	\caption{
		The original causal graph of RE models (left) together with its counterfactual alternatives for the entity bias (right).
		The shading indicates the mask of corresponding variables.
		\label{fig:graph}}
\end{figure}

We follow the prior work \cite{wang2022should} to analyze the entity bias based on causal inference.
\cite{wang2022should} builds the causal graph of RE as a directed acyclic graph: $(E, X) \rightarrow Y$ in \Cref{fig:graph}.
$X$ is the input text, $E$ denotes the entity mentions, and $Y$ is the relation extraction result.
On the edges $(X, E) \rightarrow Y$, the RE model encodes $E$ and $X$ to predict the relation $Y$.

\begin{figure}[!tb]
	\centering
	\includegraphics[width=1\linewidth]{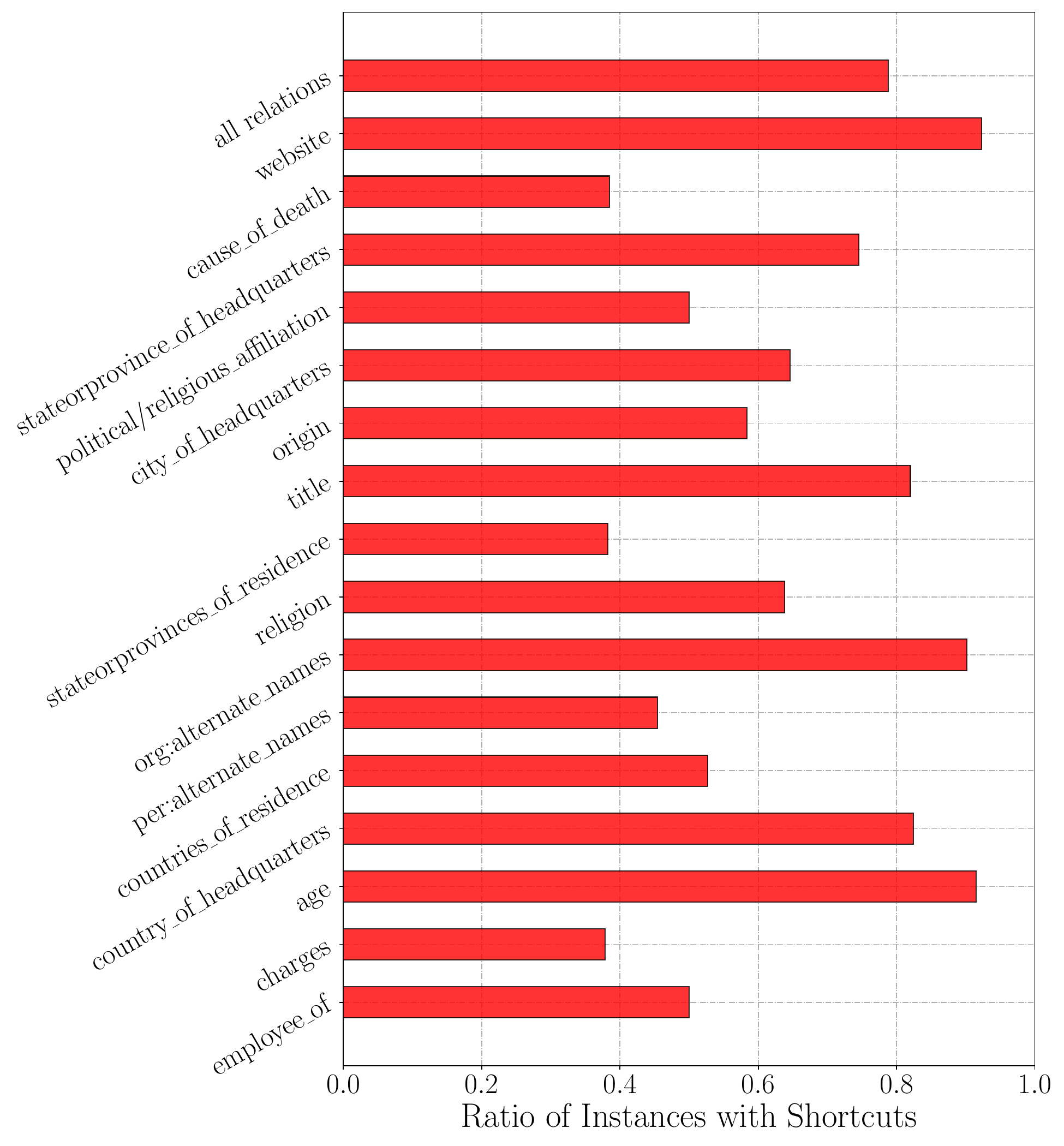}
	\caption{
	The ratio of instances with shortcuts (the entity bias is as same as the ground truth relation) in the TACRED test set.
		\label{fig:ratio_name}}
\end{figure}

Based on the causal graph displayed in \Cref{fig:graph}, we can diagnose 
whether the entities 
have shortcuts to relation.
\citet{wang2022should} distill the entity bias by counterfactual analysis, which assigns the hypothetical combination of values to variables in a way that is counter to the empirical evidence obtained from data.
We mask the tokens in $X$ to conduct the intervention $X = \bar{x}$ on $X$, while keeping the variable $E$ as the original entity mentions $e$.
In this way, the textual context is removed and the entity information is maintained.
Accordingly, the counterfactual prediction is denoted as $Y_{\bar{x}, e}$ (see \Cref{fig:graph}).
$Y_{\bar{x}, e}$ refers to the output, i.e., a probability distribution or a logit vector, where only the entity mentions are given.

\subsection{Shortcuts to the Ground-Truth Relations}
Existing work has found that the popular RE benchmarks' test sets provide abundant shortcuts from entity names to ground-truth relations \cite{wang2022should,peng2020learning}.
In other words, on many instances, the model need not ``extract'' the relation from the textual context but can infer the correct prediction directly through shortcuts from entities.

To verify these observations, we conduct a preliminary study of the shortcuts using the strong RE model LUKE \cite{yamada-etal-2020-luke} on the TACRED dataset.
We first compute the instance-wise relation extraction result in the TACRED's test set.
Then, we analyze the shortcuts from entity names to the relations based on causal inference (see details in Sec.~\ref{sec:3_1}).
We find that there exists a large portion of instances having shortcuts from entity names to the ground-truth relations.
We visualize the ratio of instances 
that present shortcuts in different relations in Fig.~\ref{fig:ratio_name}.
Last but not least, we observe similar phenomena on other models and TACREV, Re-TACRED datasets as well. 

The analyses suggest that these benchmarks do not accurately evaluate the ``extraction'' capability of RE models without the shortcuts from entity names.
In other words, these popular benchmarks are not challenging enough to evaluate whether the RE models can extract the correct relations from the textual context.
In our work, we replace the entity names to reduce the shortcuts, to mitigate the possibility that RE models rely on the shortcut of entity bias to achieve over-optimistically high RE performance.
Our \modelname is able to better simulate real-world scenarios with fewer shortcuts and higher entity diversity, which provides a better evaluation of the generalization of RE models.

\section{Entity Replacement for RE}
We present \modelname: a simple yet effective procedure to generate high-quality RE instances with entity replacements.
\modelname replaces entity names in the RE instances in a random and type-constrained way. 
We apply \modelname to the test set of TACRED to evaluate the state-of-the-art RE models' robustness under entity replacements.

\subsection{Targetting the Entities for Replacements}

We desire entity replacements to not affect the soundness of language.
As we have analyzed in Sec.~\ref{sec:2_1}, there exists a significant amount of incorrect entity annotations in TACRED.
To handle these incorrect entity annotations, we use a BERT based NER model \cite{devlin-etal-2019-bert} to re-annotate the entities in the TACRED test set.
Then, we further conduct a manual investigation over the entity annotations.
We filter out incorrectly annotated instances and only replace the named entities.
This prevents our entity name replacements from altering the ground-truth relation labels.

Besides the incorrect entity annotations,  there are also some entities 
for which replacement may inevitably cause noise.
For example, some entities belong to the \texttt{[MISC]} (miscellaneous) class.
If we replace a \texttt{[MISC]} entity with another \texttt{[MISC]} one, it is likely that we will break the semantics of the original sentence.
In contrast, replacing the \texttt{[PERSON]} and \texttt{[ORGANIZATION]} entities with those belonging to the same type generally do not affect the ground-truth relations.
We notice that all the instances in TACRED have a \texttt{[PERSON]} or \texttt{[ORGANIZATION]} entity as the subject or object.
Therefore, in our work, we focus on replacing the \texttt{[PERSON]} and \texttt{[ORGANIZATION]} entities.

\subsection{Large Lexicon of Entities}\label{sec:3_3}
We propose the following standards for selecting the new entity names for replacements:
\begin{enumerate}[leftmargin=1em]
    \item The new entity belongs to the same type as the replaced one.
    \item The new entity exists in the real world.
    \item The new entity names are more diverse.
\end{enumerate}
These three standards contribute to making the resulting instances \textit{natural} -- i.e., containing real, valid entities that are of the same class as the original entities, and are linguistically sound;
\textit{challenging} -- i.e., the new entities may not offer shortcuts to the model, which cannot easily get the correct extraction result by seeing only the entity names and \textit{comprehensive} -- i.e., the robustness of RE is evaluated on a 
more diverse set of entities.

To satisfy the above standards, we first build up a large entity name lexicon to provide the new entity names for replacements.
The size of the entity lexicon determines the diversity of entity names in our new RE benchmark \dataname.
Also, a larger entity name lexicon can help us to evaluate the generalization of RE models on more out-of-domain entity names in test time.
Therefore, in addition to the entity names appearing in the TACRED, we collect the entity names from Wikipedia belonging to the category of person and organization to enrich the entity name corpus.
Overall, we collect 24,933 organization and 902,007 person names from Wikipedia\footnote{\url{https://dumps.wikimedia.org/enwiki/latest/enwiki-latest-pages-articles.xml.bz2}} to build a large entity lexicon.

\subsection{Entity Replacements}

Based on the constructed entity lexicon, we propose ENTRE: a type-constrained and random entity replacement method.
\textbf{Type-constrained} means we replace the named entity in the type \texttt{[PERSON]} or \texttt{[ORGANIZATION]} with the new entity belonging to the same type as the original entity.
\textbf{Random} means we randomly select the entity names from our entity lexicon that consists of 24,933 organizations and 902,007 person entities for replacements.
These two principles guarantee the effectiveness of entity replacement to produce valid RE instances. We iterate over TACRED instances and replace the entity names.
We summarize \modelname as the following pipeline:
\begin{enumerate}[leftmargin=1em]
    \item Collecting the instances with predictions as same as the ground-truth relation.
    \item Replace the entity names for the collected entities in Step 1. Return to step 1.
\end{enumerate}

The above steps can be repeated for many times, and a higher repetition time leads to a higher level of the adversary.
We can stop the repeating until all the entities in the lexicon have been used.
But that will induce too long running time.
Therefore, in our work, we set the maximum number of repetitions as 200 by default.

Step 1 requires the inference on many test instances, which is time-consuming.
Considering that the F1 score's calculation of RE takes the ``no\_relation'' as the background class, we can alternatively collect the instances not belonging to the ``no\_relation'' class in Step 1.
We denote such an alternate as \modelname-fast, which saves 90\% evaluation time in the experiments.

\begin{table}[tb!]
	\centering
		
		\begin{tabular}{@{}l| c | c @{}}
			\toprule
			\textbf{Benchmark}
			& TACRED
			& \dataname \\
			\midrule
			\midrule
			$\#$ Sentences	& 15,509 & 12,419 \\
			$\#$ Tokens & 539,306 & 457,121 \\
			\bottomrule
		\end{tabular}
		
		\caption{Statistics of the TACRED and  \dataname benchmarks.}	\label{tab:data}
\end{table}

Both the \modelname and \modelname-fast are dataset-agnostic and model-agnostics.
In other words, we can apply \modelname and \modelname-fast to many RE datasets to evaluate any RE model.
In this work, to enable the easy use of \modelname, we create the challenging RE benchmark \dataname by applying \modelname on the test set of TACRED.
The overall statistics of \dataname are shown in Table 3, alongside the statistics of the original TACRED dataset.
The number of sentences in \dataname is slightly smaller than that in TACRED because we filter out the incorrectly annotated instances.
We showcase \modelname using TACRED in this paper because of its popularity on evaluating RE models and comprehensive relation-type coverage.
However, our \modelname can be applied to other RE datasets.

\begin{table*}[tb!]
	\centering
 	\begin{adjustbox}{width=\linewidth}

		\begin{tabular}{@{}l c c c @{}}
			\toprule
			\textbf{Method}
			& \textbf{TACRED}
			& \textbf{TACRED w/ \modelname (Ours)} 
            & $\Delta$
            \\
			\midrule
			\midrule
			LUKE \cite{yamada-etal-2020-luke} & 72.7 & 45.0 & $\downarrow44\%$ \\
			\midrule
			w/ Resample \cite{burnaev2015influence} & 73.1 & 45.8 & $\downarrow37\%$ \\
			w/ Entity Mask (w/o name, w/o type) \cite{zhang2017position} & 21.3 & 21.0 & $\downarrow1\%$ \\
			w/ Entity Mask (w/o name, w/ type) \cite{zhang2017position} & 44.9 & 45.9 & $\uparrow2\%$ \\
			w/ Entity Mask (w/ name, w/ type) \cite{zhang2017position} & 72.3 & 61.2 & $\downarrow15\%$ \\
   		w/ Focal \cite{lin2017focal} & 72.9 & 47.1 & $\downarrow35\%$ \\
   		w/ CoRE \cite{wang2022should} & \textbf{74.6} & \textbf{61.7} & $\downarrow17\%$ \\
			\midrule
                \midrule
			  IRE \cite{zhou2021improved} & 74.6 & 49.3 & $\downarrow34\%$ \\
			\midrule
			w/ Resample \cite{burnaev2015influence} & 73.9 & 49.6 & $\downarrow33\%$ \\
			w/ Entity Mask (w/o name, w/o type) \cite{zhang2017position} & 22.0 & 21.8 & $\downarrow1\%$ \\
			w/ Entity Mask (w/o name, w/ type) \cite{zhang2017position} & 60.9 & 61.3 & $\uparrow1\%$ \\
			w/ Entity Mask (w/ name, w/ type) \cite{zhang2017position} & 74.6 & 49.3 & $\downarrow34\%$ \\
   		w/ Focal \cite{lin2017focal} & 74.1 & 49.5 & $\downarrow32\%$ \\
   		w/ CoRE \cite{wang2022should} & \textbf{74.7} & \textbf{64.2} & $\downarrow14\%$ \\
     \bottomrule
		\end{tabular}
  	\end{adjustbox}

	\caption{F1 scores (\%) and the performance dropping of RE on the test sets of TACRED and our \dataname. The best results in each column are highlighted in \textbf{bold} font. We additionally report the performance drop (\%) compared with the performance on the original TACRED dataset.}	\label{tab:main}
\end{table*}

\section{Experiments}
In this section, we investigate \modelname and use it to evaluate the robustness of the strong RE models LUKE \cite{yamada-etal-2020-luke}, IRE \cite{zhou2021improved}, and other methods that can improve the robustness of RE. 
Our experimental settings closely follow those of previous work \cite{zhang2017position,zhou2021improved,nan2021uncovering} to ensure a fair comparison.
We organize our results and analysis as four main research questions and their answers.

\finding{Q1: How robust is relation extraction?}

\paragraph{Main Results}
We evaluate the robustness of the state-of-the-art RE models LUKE \cite{yamada-etal-2020-luke} and IRE \cite{zhou2021improved} under entity replacements.
Our experimental settings closely follow those of previous work \cite{zhang2017position,zhou2021improved,nan2021uncovering} to ensure a fair comparison.
We visualize the empirical results in Fig. \ref{fig:replace}.
We observe that the 30\% - 50\% drops in terms of F1 scores happen on the state-of-the-art RE models after entity replacements.
These results suggest that there remains a large gap between the current research and the really effective RE models robust to entity replacements. 

We compare the F1 scores on TACRED and \dataname, the challenging RE benchmark produced by our \modelname, in \Cref{tab:main}.
We can see that the state-of-the-art LUKE has a significant performance drop in our challenging \dataname; there is a 44\% relative decrease (in the models’ F1) in \dataname as
compared to their results before entity replacements.

\begin{table*}[!ht]
	\centering
	\renewcommand\arraystretch{2.9}
	\begin{adjustbox}{width=\linewidth}
		
		\begin{tabular}{@{}l|c|c|c @{}}
			\toprule 
			\textbf{Original Instance}
			& \textbf{Original Prediction}
			& \textbf{New Entity Names}
			& \textbf{New Prediction}
			\\ \midrule
			\multirow{2}{7.5cm}{Finance Ministry spokesperson Chileshe Kandeta who confirmed this on Sunday said Magande signed a loan agreement of 31 million dollars with the \underline{ADF} for the country's \uwave{Poverty Reduction Budget Support}.} & 
			\multirow{2}{*}{no\_relation \cmark}&
			\multirow{2}{4cm}{\underline{American Association of} \underline{University Women}, \uwave{Willingboro Chapter}}&
			\multirow{2}{*}{members \xmark}
   
			\\
			&&&
			\\
			\midrule
			
			\multirow{2}{7.5cm}{John Graham, a 55-year-old man from Canada, is accused of shooting \underline{Aquash} in the head and leaving her to die on the Pine Ridge reservation in \uwave{South Dakota}.}
			& 
			
			\multirow{2}{*}{stateorprovince\_of\_death \cmark}&
			
			\multirow{2}{4cm}{\underline{Liu Shaozhuo},\ \ \ \ \ \ \ \ \ \ \  \uwave{South Dakota}}&
			
			\multirow{2}{*}{no\_relation \xmark}
			
			\\
			&&&
			\\
			\midrule

			\multirow{2}{7.5cm}{After the staffing firm \underline{Hollister Inc} lost 20 of its \uwave{85} employees, it gave up nearly a third of its 3,750-square-foot Burlington office, allowing the property owner to put up a dividing wall to create a space for another tenant.}
			& 
			
			\multirow{2}{*}{number\_of\_employees/members \cmark}&
			
			\multirow{2}{4cm}{\underline{Yoruba Academy}, \uwave{85}}&
			
			\multirow{2}{*}{alternate\_names \xmark}
			
			\\
			&&&
			\\
			\midrule

			\multirow{2}{7.5cm}{\uwave{Kercher} 's mother, \underline{Arline Kercher}, tells court in emotional testimony that she will never get over her daughter 's brutal death.}
			& 
			
			\multirow{2}{*}{children \cmark}&
			
			\multirow{2}{4cm}{\underline{Sanju Yadav}, \uwave{Matti Koistinen}}&
			
			\multirow{2}{*}{no\_relation\xmark}
			
			\\
			&&&
			\\
			\midrule

			\multirow{2}{7.5cm}{Lt. \uwave{Assaf Ramon}, the son of Israel's first astronaut, Col. \underline{Ilan Ramon}, who died in the space shuttle Columbia disaster in 2003, was killed Sunday when an F16-A plane he was piloting crashed in the hills south of Hebron in the West Bank.}
			& 
			
			\multirow{2}{*}{children \cmark}&
			
			\multirow{2}{4cm}{\underline{Aaron Morgan}, \ \ \ \ \ \ 
            \ \ \uwave{Ángel Guillermo Heredia Hernández}}&
			
			\multirow{2}{*}{no\_relation \xmark}
			
			\\
			&&&
			\\
			\midrule

			\multirow{2}{7.5cm}{Police have released scant information about the killing of 61-year-old \underline{Carol Daniels}, whose body was found Sunday inside the Christ Holy Sanctified Church, a weather-beaten building on a rundown block near downtown Anadarko in southwest \uwave{Oklahoma}.}
			& 
			
			\multirow{2}{*}{stateorprovince\_of\_death \cmark}&
			
			\multirow{2}{4cm}{\underline{Mao Weiming}, \uwave{Oklahoma}}&
			
			\multirow{2}{*}{no\_relation \xmark}
			
			\\
			&&&
			\\

			\bottomrule
		\end{tabular}
	\end{adjustbox}
	\caption{A case study for LUKE on the relation extraction benchmark TACRED and our \dataname.
		\underline{Underlines} and \uwave{wavy lines} highlight the subject and object entities respectively.
		We report the original prediction, the new entity names for replacements and the prediction in \dataname.}
	\label{tab:case}
\end{table*}

\paragraph{Case Study}
We conduct case studies to empirically examine the effects of our entity replacements of \modelname.
\Cref{tab:case} gives a qualitative comparison example between the RE results on TACRED and our \dataname. 
The results show that our \modelname misleads the strong RE model LUKE to predict incorrect relations.
For example, given the TACRED instance ``\textit{Finance Ministry spokesperson Chileshe Kandeta who confirmed this on Sunday said Magande signed a loan agreement of 31 million dollars with the \underline{ADF} for the country 's \uwave{Poverty Reduction Budget Support}.}'', there is no relation between the subject and object existing in the text.
After the entity replacement, LUKE believes that the relation between them is ``\textit{members}''. 

The entity bias can account for this result, where given only the entity mentions \textit{American Association of University Women} and \textit{Willingboro Chapter}, the RE model returns the relation ``\textit{members}'' without any textual context. 
This implies that the model makes the prediction for the original input relying on the entity mentions, which leads to the wrong RE prediction.
In our work, we replace the original entities with the new ones that convey the entity bias different from the ground-truth label to test the generalization of RE models under entity bias.


\paragraph{Memorizing or Reasoning?}
We propose \modelname to test the ability to use the textual context to infer the relations. 
As the entity replacements of \modelname do not affect the ground-truth relations, RE models should be robust against entity name changes. 
However, we observe the large performance drops from our entity replacements. 

Therefore, we conclude that the strong RE model LUKE is apt to memorize the entity name patterns for predicting relations and is more brittle when the entities that convey the biases are different from the ground-truth relations existing in the input text.
To make RE models more robust, we believe an important future direction is to develop context-based reasoning approaches, taking advantage of inductive biases on the textual context that determines the relations.

\begin{figure}[!tb]
	\centering
	\includegraphics[width=1\linewidth]{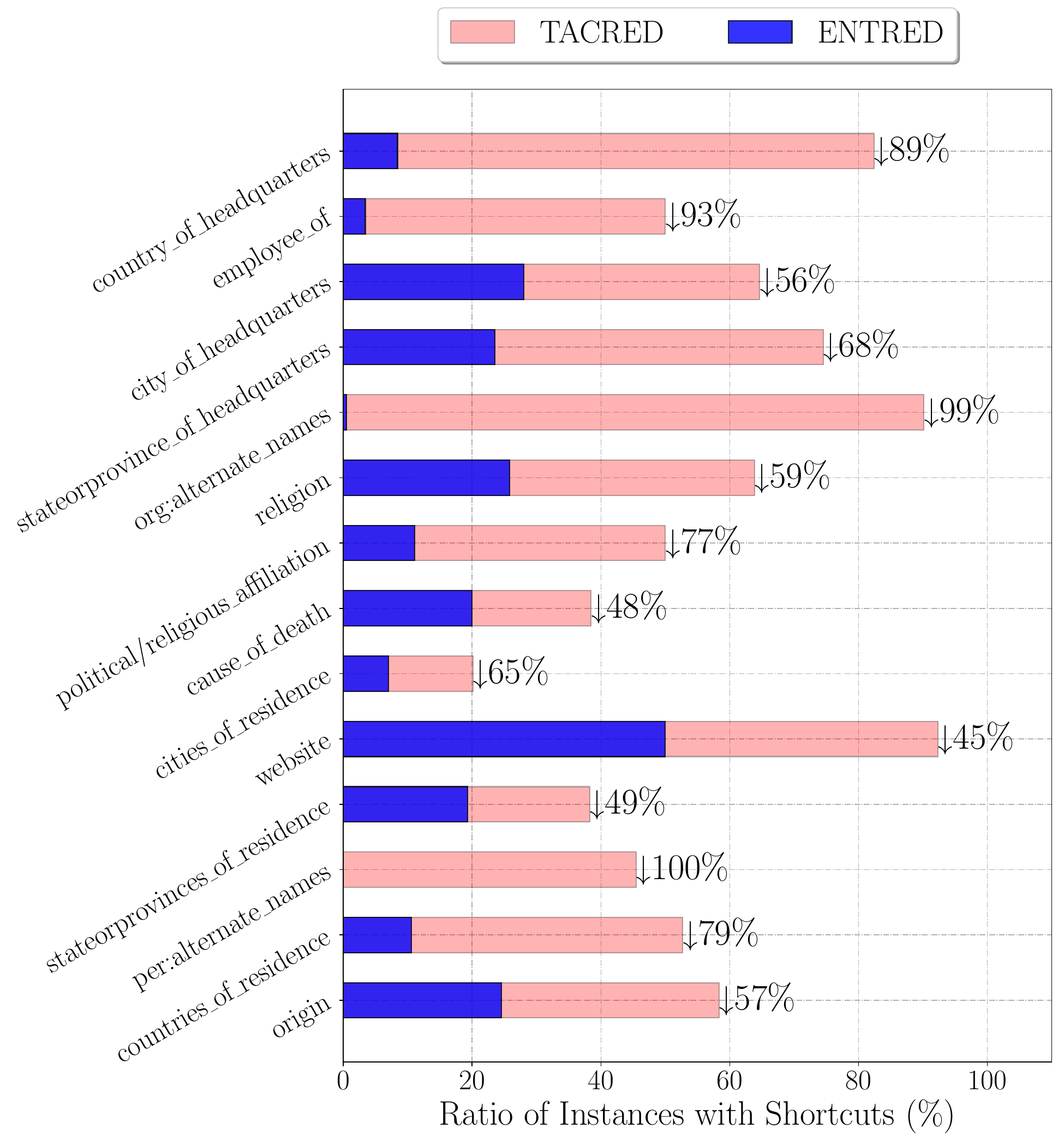}
	\caption{
	\modelname significantly reduces the ratio of instances with shortcuts (the entity bias is as same as the ground truth relation) compared with TACRED.
		\label{fig:compare_shortcut}}
\end{figure}

\finding{Q2: Does \modelname reduce shortcuts?}

\paragraph{\modelname leads to fewer shortcuts from entity names to ground-truth relations} 
We perform causal inference over \dataname to analyze how many instances have shortcuts from entity names to the ground-truth relations after the entity replacements.
We present the comparison of the shortcut ratio on \dataname and TACRED on different relations in Fig.~\ref{fig:compare_shortcut}. 
We observe that \dataname greatly reduces the shortcuts for more than 50\% instances on most relations.
As a result, when being evaluated using \dataname, RE models have to extract the informative signals describing the ground-truth relations from the textual context, rather than rely on the shortcuts from the entity names.

\finding{Q3: Does \modelname improve diversity?}

\begin{figure}[!tb]
	\centering
	\includegraphics[width=1\linewidth]{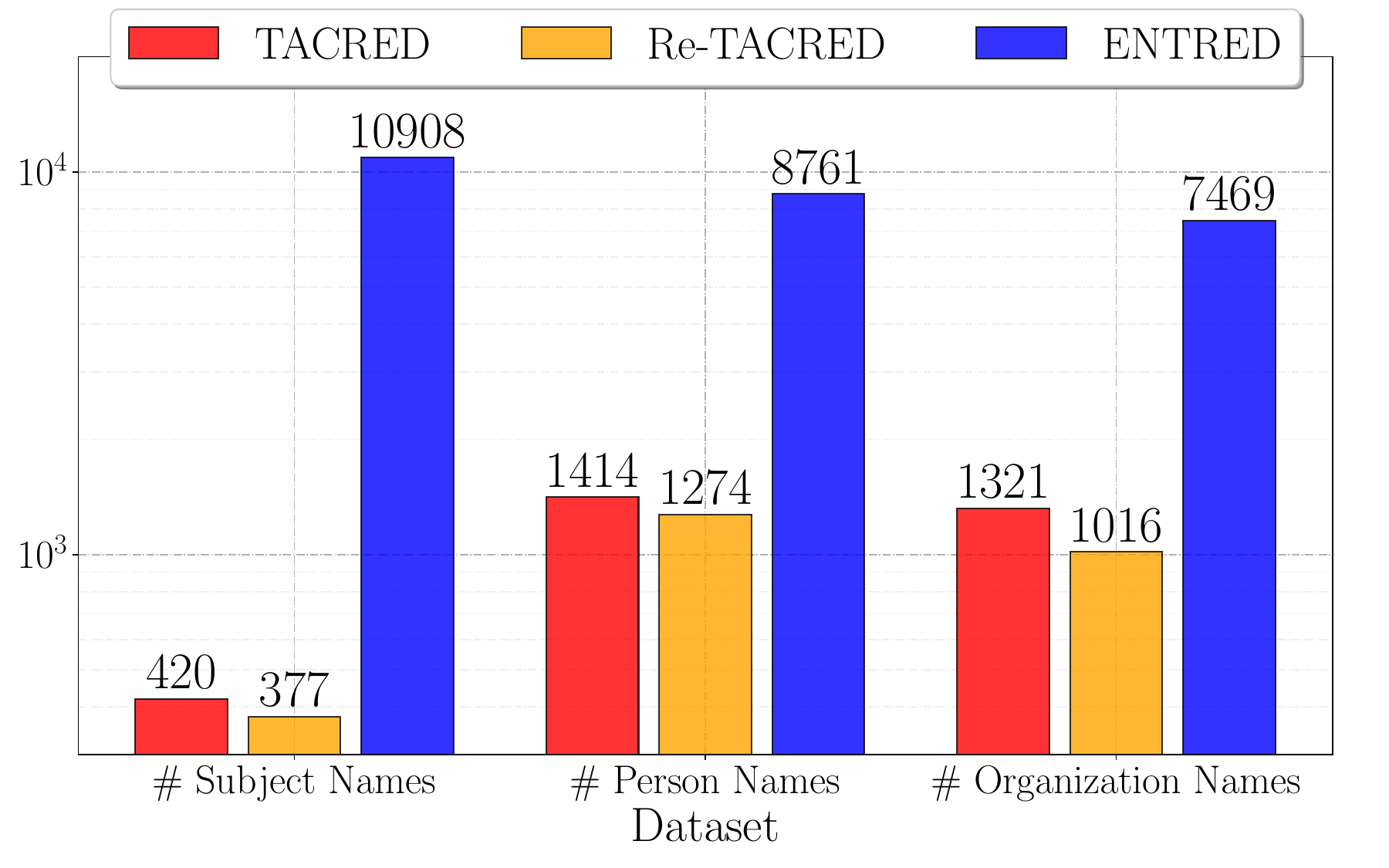}
	\caption{
	The number of subject entity names, person entity names, and organization entity names in the test set of TACRED (red) and \dataname (blue).
		\label{fig:number_name_new}}
\end{figure}

\paragraph{Comparison between \dataname and existing benchmarks.} As we have analyzed in Sec.~\ref{sec:2_1}, the diversity of entity names in the existing benchmarks TACRED, TACREV and Re-TACRED are rather limited.
These limitations hinder the evaluation of the generalization and generalization 
of RE.
In our work, thanks to our larger lexicon built from the Wikipedia entity names, our \dataname have much higher diversity than the TACRED and Re-TACRED, as shown in Fig.~\ref{fig:number_name_new}.
With these diverse entity names, \dataname is able to evaluate the performance of RE models on a larger scale of diverse entities, which better imitates the real–world scenario.

\finding{Q4: How to improve the generalization?}

\paragraph{Methods} 
In our work, we consider the following methods to improve the generalization of RE: 
(1) \textbf{Focal} \cite{lin2017focal} adaptively reweights the losses of different instances so as to focus on the hard ones. 
(2) \textbf{Resample} \cite{burnaev2015influence} up-samples rare categories by the inversed sample fraction during training.
(3) \textbf{Entity Mask} \cite{zhang2017position}: masks the entity mentions with special tokens to reduce the over-fitting on entities.
(4) \textbf{CoRE} \cite{wang2022should} is a causal inference based method that mitigates entity bias.

\paragraph{Results \& Analysis}
The results of the above methods on the RE model are shown in \Cref{tab:main}.
The recently proposed causal inference based debiasing method CoRE offers the best improvements against our entity replacements (($45.0\% \rightarrow 61.7\%$)).
We conjecture that this is because it mitigates the biasing signals from entity names, which enhances its entity-level generalization ability and makes RE models focus more on the textual context for inference, resulting in a better generalization under entity name replacements.
Other methods, however, lead to lower improvements for LUKE, potentially because they cannot effectively capture the biased patterns between relations and entity names.

\section{Related Work}
Relation extraction (RE) is a sub-task of information extraction that aims to identify semantic relations between entities from natural language text \cite{zhang2017position}. 
RE is the key component for building relation knowledge graphs, and it is of crucial significance to natural language processing applications such as structured search, sentiment analysis, question answering, and summarization \cite{huang2017deep}.
Early research efforts \cite{nguyen-grishman-2015-relation,wang-etal-2016-relation,zhang2017position} train RE models from scratch based on lexicon-level features.
The recent RE work fine-tunes pretrained language models (PLMs; \citealt{devlin-etal-2019-bert,liu2019roberta}).
For example, K-Adapter~\cite{wang2020k} fixes the parameters of the PLM and uses feature adapters to infuse factual and linguistic knowledge.
Recent work focuses on utilizing the entity information for RE \cite{zhou2021improved,yamada-etal-2020-luke}, but this leaks superficial and spurious clues about the relations \cite{zhang2018graph}.
Despite the biases in existing RE models, scarce work has discussed the spurious correlation between entity mentions and relations that cause such biases.
Our work builds an automated pipeline to generate natural instances with fewer shortcuts and larger coverage at scale to reflect the serious effects of entity bias on the RE models.

There is also work in other domains aiming to evaluate models’ generalization to perturbed inputs.
For example, \citet{jia2017adversarial} attacks reading comprehension models by adding word sequences to the input. 
\citet{gan2019improving} and \citet{iyyer-etal-2018-adversarial} paraphrase the input to test models’ over-sensitivity. 
\citet{jones-etal-2020-robust} target adversarial typos. 
\citet{si2020benchmarking} propose a benchmark for reading comprehension with diverse types of test-time perturbation. 
These works focus on different domains than our research does, and they do not consider the composition of RE examples. 
Little attention is drawn to the entities in the sentences, and many attacks (e.g. character swapping, word injection) may make the perturbed sentences invalid. 
To the best of our knowledge, this work is among the first to propose a straightforward, dedicated pipeline for generating natural adversarial examples for the RE task, which takes into account the serious effects of entity bias in RE models.

\section{Conclusion}
Our contributions in this paper are three-fold.
1) Methodology-wise: we propose \modelname, an end-to-end entity replacement method that reduces the shortcuts from entity names to ground-truth relations.
2) Resource-wise: we develop \dataname, a straightforward method for generating natural and counterfactual entity replacements for RE, which produces \dataname, a benchmark for auditing the generalization of RE models under entity bias. 
3) Evaluation-wise: our experimental results and analysis provide answers to four main research questions on the generalization of RE.
We believe \dataname and the entity replacement method \modelname can benefit the community working to increase the RE models' generalization under entity bias.   

\section*{Acknowledgement}
The authors would like to thank the anonymous reviewers for their discussion and feedback.

Wenxuan Zhou and Muhao Chen are supported by the NSF Grant IIS 2105329, the NSF Grant ITE 2333736, 
the DARPA MCS program under Contract No. N660011924033 with
the United States Office Of Naval Research, a Cisco Research Award, two Amazon Research Awards, and a Keston Research Award.
Fei Wang is supported by the Annenberg Fellowship and the Amazon ML Fellowship.
Yiwei Wang and Bryan Hooi are supported by NUS ODPRT Grant A-0008067-00-00, NUS ODPRT Grant R252-000-A81-133, and Singapore Ministry of Education Academic Research Fund Tier 3 under MOEs official grant number MOE2017-T3-1-007.

\bibliography{acl2020}
\bibliographystyle{acl_natbib}

\end{document}